# MULTIPROCESSOR SCHEDULING FOR TASKS WITH PRIORITY USING GA

Mrs.S.R.Vijayalakshmi
Lecturer, School of Information Technology and Science,
Dr.G.R.D College of Science,
Coimbatore -14, India.
.

Dr.G.Padmavathi
Professor and Head, Dept.of Computer Science,
Avinashilingam University for Women,
Coimbatore – 43,India.
.

**Abstract** - *Multiprocessors have emerged as a powerful computing means for running real-time applications, especially where a uni-processor system would not be sufficient enough to execute all the tasks. The high performance and reliability of multiprocessors have made them a powerful computing resource. Such computing environment requires an efficient algorithm to determine when and on which processor a given task should execute. In multiprocessor systems, an efficient scheduling of a parallel program onto the processors that minimizes the entire execution time is vital for achieving a high performance. This scheduling problem is known to be NP- Hard. In multiprocessor scheduling problem, a given program is to be scheduled in a given multiprocessor system such that the program's execution time is minimized. The last job must be completed as early as possible. Genetic algorithm (GA) is one of the widely used techniques for constrained optimization problems. Genetic algorithms are basically search algorithms based on the mechanics of natural selection and natural genesis. The main goal behind research on genetic algorithms is robustness i.e. balance between efficiency and efficacy. This paper proposes Genetic algorithm to solve scheduling problem of multiprocessors that minimizes the make span.*

*Ke*y-*Words: -* **Task Scheduling, Genetic Algorithm (GA), parallel processing.**

## I INTRODUCTION

Real-time systems are software systems in which the time at which the result is produced is as important as the logical correctness of the result. That is, the quality of service provided by the real-time computing system is assessed based on the main constraint 'time'. Real-time applications span a large range of activities, which include production automation, embedded systems, telecommunication systems, nuclear plant supervision, surgical operation monitoring, scientific experiments, robotics and banking transactions.

Scheduling is an important aspect in real-time systems to ensure soft and hard timing constraints. Scheduling tasks involves the allotment of resources and time to tasks, to satisfy certain performance needs. In a real-time application, real-time tasks are the basic executable entities that are scheduled. The tasks may be periodic or aperiodic and may have soft or hard real-time constraints. Scheduling a task set consists of planning the order of execution of task requests so that the timing constraints are met. Multiprocessors have emerged as a powerful computing means for running real-time applications, especially where a uni-processor system would not be sufficient enough to execute all the tasks by their deadlines. The high performance and reliability of multiprocessors have made them a powerful computing means in time-critical applications.

Real-time systems make use of scheduling algorithms to maximize the number of real-time tasks that can be processed without violating timing constraints. A scheduling algorithm provides a schedule for a task set that assigns tasks to processors and provides an ordered list of tasks. The schedule is said to be feasible if the timing constraints of all the tasks are met. All scheduling algorithms face the challenge of creating a feasible schedule. The two main objectives of task scheduling in real-time systems are meeting deadlines and achieving high resource utilization. Section 1 deals with Introduction. Section 2 deals about the task scheduling in parallel systems. Section 3 about the back ground study. Section 4 about the Genetic algorithm. Section 5 about algorithm design. Section 6 about the experimental results and final section gives the conclusion.

## II TASK SCHEDULING IN PARALLEL SYSTEMS

Multiprocessor scheduling problems can be classified into many different classes based on characteristics of the program and tasks to be scheduled, the multiprocessor system, and the availability of information. A deterministic scheduling problem is one in which all information about the tasks and their relations to each other, such as execution time and precedence relations are known to the scheduling algorithm in advance. Such problems, also known as static scheduling problems. In contrast to nondeterministic scheduling problems in which some information about tasks and their relations may be non-determinable until runtime, i.e., task execution time and precedence relations may be determined by data input.

Within the class of deterministic scheduling problems, the following are the constraints:
1. The number of tasks.
2. Execution time of the tasks.
3. Precedence of the tasks.
4. Topology of the representative task graph.
5. Number of processors.
6. Processors uniformity.
7. Inter task communication.
8. Performance criteria.

## 2.1 Problem description
The goal of multiprocessor scheduling is to find an optimization solution to minimize the overall execution time for a collection of subtasks that compete for computation.
Given material for problem
- A multiprocessor system with 'm' machines.
- A task represented by a DAG.
- The estimated execution duration of every subtask.

## 2.2 Problem Statement
The goal of multiprocessor scheduling is to find an optimization algorithm to minimize the overall execution time for a collection of subtasks that compete for computation and also the maximum utilization of the processor time.

A (homogeneous multiprocessor system is composed of a set P= {P1, …Pm} of 'm' identical processors. They are connected by a complete communication network where all links are identical. Task preemption is not allowed. While computing, processor can communicate through one or several of its links.

A schedule to,
$$\text{Minimize} \left\{ \underset{j=1}{\overset{n}{\text{Max}}} \left[ \text{finish time}(V_j) \right] \right\}$$

where, the schedule determines, for each subtask, both the processor on which execution will take place and the time interval within which it will be executed. The Problem statement can be given as follows:
"Schedule '*n*' jobs to '*m*' processors such that the maximum span is minimized".

## 2.3 Model
A parallel program can be represented as a directed acyclic graph (DAG), G = (V,E), where V is the set of nodes each of which represents a component subtask of the program and E is the set of directed edges that specify both precedence constraints and communication paths among nodes. In the DAG model, each node label gives the execution time for the corresponding task. A task cannot start until all of its predecessor tasks are complete.

For a task graph TG = (V, E):
- Ti is a predecessor of Tj and Tj is a successor of Ti
- Ti is an ancestor of Tj and Tj is a child of Ti if there is a sequence of directed edges leading from Ti to Tj.
- PRED(Ti) – The set of predecessor of Ti
- SUCC(Ti) – The set of successor of Ti
- Et(Ti) – The execution time of Ti.

A simple task graph TG, with 8 tasks is illustrated in fig 1.

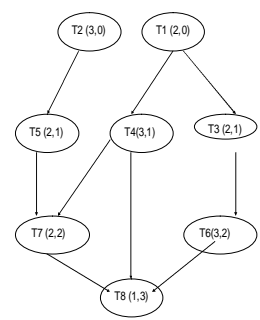

**Fig 1. A task graph TG**

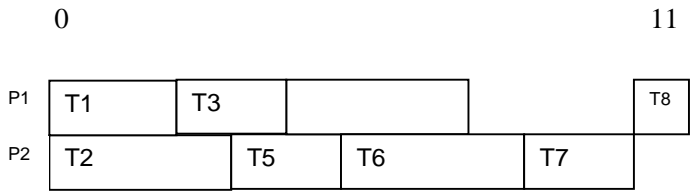

**Fig 2 Gantt chart for scheduling two tasks**

The problem of optimal scheduling a task graph onto a multiprocessor system with 'p' processors is to assign the computational tasks to the processors so that the precedence relations are maintained and all of the tasks are completed in the shortest possible time. The time that the last task is completed is called the finishing time (FT) of the schedule. Fig 2 shows a schedule for two processors displayers as Gantt chart. Fig 2 illustrates a schedule displayed as a Gantt chart for the example task graph TG using two processors. This schedule has a finishing time of 11 units of time. An important lower bound for the finishing time of any schedule is the critical path length. The critical path length, $t_{cp}$ of a task graph is defined as the minimum time required to complete all of the tasks in the task graph.

**2.4 Height of a task graph:**
The height of a task in a task graph is defined as-

0, if PRED (Ti) = 0
Height (Ti)      = 1 + max height (Tj)

This height function indirectly conveys precedence relations between the tasks. If the task Ti is an ancestor of task Tj, then height (Ti) < height (Tj). If there is no path between the two tasks, then there is no precedence relation between them and order of their execution can be arbitrary.

### III BACKGROUND

List scheduling techniques assign a priority to each task to be scheduled and then sort the list of tasks in decreasing priority. As processors become available, the highest priority task in the task list is assigned to be processed and removed from the list. If more than one task has the same priority, selection among the candidate tasks is typically random.

In order to allocate parallel applications to maximize throughput, task precedence graph (TPG) and a task interaction graph (TIG) are modeled.

The system usually schedules tasks according to their deadlines, with more urgent ones running at higher priorities. . The Earliest Deadline First (EDF) algorithm is based on the dead line time constraint. The tasks are ordered in the increasing order of their deadlines and assigned to processors considering earliest deadline first.

In multiprocessor real time systems static algorithms are used to schedule periodic tasks whose characteristics are known a priori. Scheduling aperiodic tasks whose characteristics are not known a priori requires dynamic scheduling algorithms. Some researchers analyze the task scheduling problems based on the dynamic load balancing. It minimizes the execution time of single applications running in parallel on multi computer systems. It is essential for the efficient use of highly parallel systems with non uniform problems with unpredictable load estimates. In a distributed real time systems, uneven task arrivals temporarily overload some nodes and leave others idle or under loaded.

In the proposed work, the GA technique is involved to solve the task scheduling problem.

In the proposed GA technique, the tasks are arranged as per their precedence level before applying GA operators. The cross over operator is applied for the tasks having different height and mutation operator is applied to the task having the same height. The fitness function attempts to minimize processing time.

### IV GENETIC ALGORITHMS

Genetic algorithms try to mimic the natural evolution process and generally start with an initial population of individuals, which can either be generated randomly or based on some algorithm. Each individual is an encoding of a set of parameters that uniquely identify a potential solution of the problem. In each generation, the population goes through the processes of crossover, mutation, fitness evaluation and selection. During crossover, parts of two individuals of the population are exchanged in order to create two entirely new individuals which replace the individuals from which they evolved. Each individual is selected for crossover with a probability of crossover rate. Mutation alters one or more genes in a chromosome with a probability of mutation rate.

For example, if the individual is an encoding of a schedule, two tasks are picked randomly and their positions are interchanged. A fitness function calculates the fitness of each individual, i.e., it decides how good a particular solution is. In the selection process, each individual of the current population is selected into the new population with a probability proportional to its fitness. The selection process ensures that individuals with higher fitness values have a higher probability to be carried onto the next generation, and the individuals with lower fitness values are dropped out. The new population created in the above manner constitutes the next generation, and the whole process is terminated either after a fixed number of generations or when a stopping criteria is met.

The population after a large number of generations is very likely to have individuals with very high fitness values, which imply that the solution represented by the individual is good; it is very likely to achieve an acceptable solution to the problem. The population size, the number of generations, the probabilities of mutation and crossover are some of the other parameters that can be varied to obtain a different genetic algorithm.

## V ALGORITHM DESIGN

For genetic algorithm, a randomly generated initial population of search nodes is required. It impose the following height ordering condition on schedules generated:
"The list of tasks within each processor of schedule is ordered in an ascending order of their height".

### 5.1 Initial Population
**Algorithm to generate initial population**
*{Generates a schedule of task graph TG for multiprocessor*
*system with p processors}*
1. [Initialize] Compute height for every task in TG
2. [Separate tasks according to their height]
3. [loop p-1 times] For each of first p-1 processors, do step 4
4. [form the schedule for a processor]
5. [Last processor] Assign remaining tasks in the set to last processor.

### 5.2 Fitness Function
Multiprocessor scheduling problem will also consider factors such as throughput, finishing time and processor utilization.

Genetic algorithm is based on finishing time of a schedule. The finishing time of a schedule, S is defined as follows :

$$FT(S) = \max_{P_j} ftp(P_j)$$

where, $ftp(P_j)$ is the finishing time for the last task in processor $P_j$. To maximize the fitness function, one need to convert the finishing time into maximization form. This can be done by defining the fitness value of schedule, S, as follows :

$$C_{max} - FT(S)$$

Where, $C_{max}$ is the maximum finishing time observed so far. Thus the optimal schedule would be the smallest finishing time and a fitness value larger than the other schedules.

### 5.3 Genetic Operators
Function of genetic operators is to create new search nodes based on the current population of search nodes. By combining good structures of two search nodes, it may result in an even better one. For multiprocessing scheduling problem, the genetic operators used must enforce the intra processor precedence relations, as well as completeness and uniqueness of the tasks in the schedule. For multiprocessor scheduling, certain portions of the schedule may belong to the optimal schedule. By combining several of these optimal parts, one can find the optimal schedule efficiently. For multiprocessing scheduling problem, the genetic operators used must enforce the intraprocessor precedence relations, as well as completeness and uniqueness of the tasks in the schedule.

#### 5.3.1 Crossover
The new strings can be created by exchanging portions of two strings using following method:
1. Select sites which differ in height where the lists can be cut into two halves
2. Exchange bottom halves of P1 in string A and string B
3. Exchange bottom halves of P2 in string A and string B.

For multiprocessor scheduling, one should ensure that the precedence relation is not violated and that the completeness and uniqueness of tasks still holds after crossover

#### 5.3.2 Reproduction
Reproduction process forms a new population of strings by selecting string in the old population based on their fitness values. The selection criterion is that the strings with higher fitness value should have higher chance of surviving to next generation. Good strings have high fitness value and hence should be preserved in the next generation.

#### 5.3.3 Mutation
For multiprocessor scheduling problem, mutation is applied by randomly exchanging two tasks with same height.

### 5.4 Algorithm using GA
//Algorithm Find-Schedule

1. [Initialize]
2. Repeat steps 3 to 8 until algorithm is convergent
3. Compute fitness values for each string in the initial population
4. Perform Reproduction. Store string with highest fitness values in BEST_STRING.
5. Perform crossover
6. Perform mutation
7. Preserve the best string in BEST_STRING

## VI EXPERIMENTAL RESULTS

The genetic algorithm has been implemented and tested. The following are the assumptions and conditions under which the experiment is conducted. Assumptions about the task number range from 8 to 110. The number of successors that each task node is allowed is a random number between 3 and 6. The execution time for each task random number between 1 and 25. The task graphs are tested on a list-scheduling algorithm. The genetic algorithm used the following parameters throughout the simulation.

- Population size = 20.
- Maximum number of iterations = 500.

The simulation is performed using MATLAB.

**6.1 Comparison between GA and LSH**

This section compares the list scheduling heuristic (LSH) with the genetic algorithm (GA). List scheduling is taken because the tasks are arranged as per the precedence relations. The proposed method using GA also takes the tasks in their precedence relation sequence.

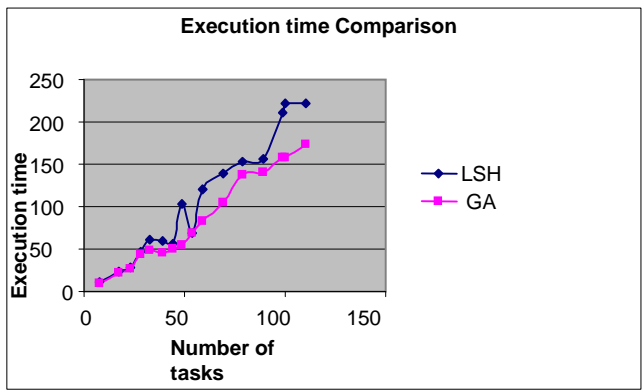

**Fig 3 Task Vs Execution Time**

From the fig 3 the LSH and GA produces almost same scheduling time when the number tasks in range 8 to 28. When the number of tasks increase GA gives the better solution. When the tasks are more than 100, the GA gives the best solution.

**Table 1: Execution time GA and LSH**

| Number of tasks | ISH Execution Time | GA Execution Time |
|---|---|---|
| 8 | 11 | 10 |
| 17 | 24 | 22 |
| 23 | 28 | 26 |
| 28 | 47 | 43 |
| 39 | 60 | 46 |
| 44 | 56 | 50 |
| 49 | 103 | 55 |
| 54 | 68 | 68 |
| 59 | 121 | 83 |
| 69 | 139 | 104 |
| 79 | 153 | 137 |
| 89 | 157 | 141 |
| 100 | 222 | 158 |

One can infer from the table 1 the finish time of all the range tasks are lesser in the case of GA when compared with LSH. When the number of tasks is increased, GA only gives the minimum finish time.

**Table 2: Execution time of algorithm**

| Number of tasks | GA time to schedule task |
|---|---|
| 8 | 3.261 |
| 17 | 3.249 |
| 28 | 2.593 |
| 33 | 3.125 |
| 39 | 2.688 |
| 44 | 2.625 |
| 49 | 2.859 |
| 54 | 3.537 |
| 59 | 0.885 |
| 69 | 0.842 |
| 79 | 3.186 |
| 89 | 3.155 |
| 99 | 3.217 |
| 100 | 2.311 |

The table 2 indicates the time taken by the GA algorithm to compute the scheduling time for tasks among many number of processors.

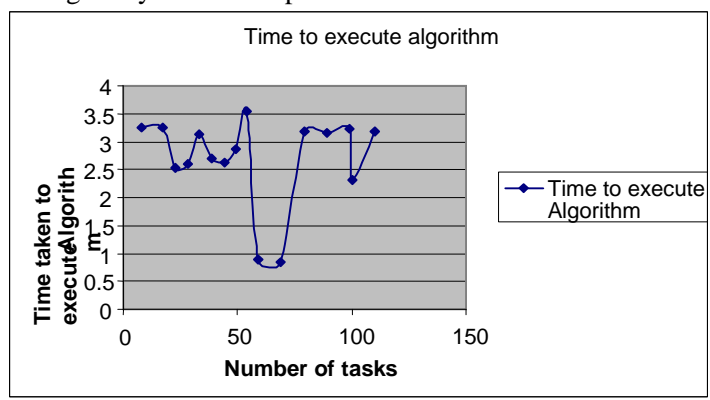

**Fig 4: Execution time of algorithm**

From the fig 4 one can infer the time required to execute the algorithm is minimum when the number of tasks range between 55 and 70. Hence the tasks in this range GA finds the maximum fitness within the minimum period of time.

Table 3 Number of processor Vs. GA and LSH

| No of processor | Best minimum time GA | LSH minimum time |
|---|---|---|
| 2 | 22 | 24 |
| 3 | 26 | 28 |
| 3 | 48 | 61 |
| 4 | 46 | 60 |
| 4 | 55 | 103 |
| 4 | 158 | 222 |
| 4 | 174 | 222 |

When the number of processors is increased the LSH takes more time to find the schedule and also the processors are not utilized to the maximum limit. GA gives best solution and the processor performance is increased and the processors are utilized to their maximum limit. The time slot of any processor is not wasted.

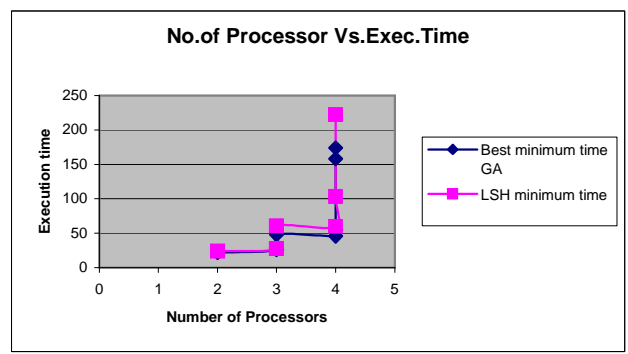

**Fig 5: Number of processor Vs. GA and LSH**

**Table 4: Precedence relation Vs. Execution time**

| Height | Best minimum time GA | LSH minimum time |
|---|---|---|
| 3 | 10 | 11 |
| 5 | 22 | 24 |
| 5 | 46 | 60 |
| 5 | 50 | 56 |
| 6 | 158 | 222 |
| 7 | 174 | 222 |

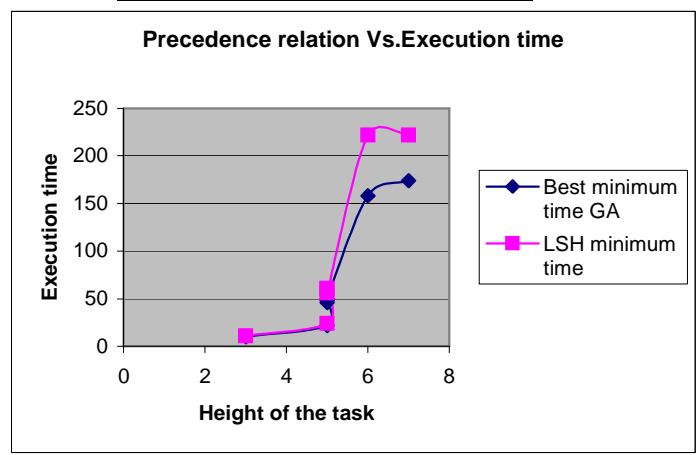

**Fig 6: Precedence relation Vs. Execution time**

One can infer from the table 4 and fig 6 the LSH and GA are producing the same result between the height 3 to 5.But when the height are more than 5 only GA produces the best result.

**Table 5 Comparisons between LSH and GA**

| Parameter | LSH | GA |
|---|---|---|
| Computation time (task) | Minimum | Higher |
| Communication time | Minimum | Higher |
| Make span | Higher | Minimum |
| Cost | Minimum | Optimum |
| Scheduling time | Higher | Minimum |
| Execution time when Problem size increases | Takes longer time to solve | Minimum |

The table 5 shows the comparison of the parameters between GA and LSH. Table 6 gives population vs. GA. In only 20 generations, the GA finds an optimal solution for all the tasks. A suitable solution for 54 tasks is found in less number of generations.

**Table 6 Population Vs. GA**

| Number of tasks | Population | optimal Schedule | GA | OS-GA/GA * 100 |
|---|---|---|---|---|
| 8 | 20 | 11 | 10 | 10.0 |
| 17 | 20 | 24 | 22 | 9.1 |
| 23 | 20 | 28 | 26 | 7.7 |
| 44 | 20 | 56 | 50 | 12.0 |
| 49 | 20 | 103 | 55 | 87.3 |
| 54 | 20 | 68 | 68 | 0.0 |
| 69 | 20 | 139 | 104 | 33.7 |
| 79 | 20 | 153 | 137 | 11.7 |
| 89 | 20 | 157 | 141 | 11.3 |
| 99 | 20 | 211 | 158 | 33.5 |

All the results show that Genetic algorithm is better than List Scheduling.

The advantages of the GA are simple to use, requires minimal problem specific information, and is able to effectively adapt in dynamically changing environments. It also indicates that the GA is able to adapt automatically to changes in the problem to be solved. Although performance decreases after a target change, the GA immediate begins to improve solutions and is ultimately able to find near optimal solutions even when one or more processors are significantly slower than normal.

**VII CONCLUSIONS**

The problem of scheduling of tasks to be executed on a multiprocessor system is one of the most challenging problems in parallel computing. Genetic algorithms are well adapted to multiprocessor scheduling problems. As the resources are increased available to the GA, it is able to find better solutions. The trade off for the increased resources used by the GA is a significantly longer execution time than traditional methods. Overall, the GA appears to be the most flexible algorithm for problems using heterogeneous processors. It also indicates that the GA is able to adapt automatically to changes in the problem to be solved. The advantages of the GA approach presented here are that it is simple to use, requires minimal problem specific information, and is able to effectively adapt in dynamically changing environments. The genetic algorithm with the combination of other scheduling technique may be applied to solve the multiprocessor scheduling problem. It may give even better performance and processor utilization. It may overcome the larger execution time of the algorithm.

IEEE transactions on parallel and distributed systems, Vol. 14, No.7, pp 686-699.

10. Rabi N.Mahapatra and Wei Zhao, July 2005, An energy efficient slack distribution technique for multimode distributed real time embedded systems, IEEE transactions on parallel and distributed systems, Vol. 16, No.7, pp 650-662.

11. Theodore P.Baker, August 2005, An analysis of EDF schedulability on a multiprocessor, IEEE transactions on parallel and distributed systems, Vol. 16, No.8, pp 760-768.

12. Eitan Frachtenberg, Fabrizio Petrini, November 2005, Adaptive parallel job scheduling with flexible coscheduling, IEEE transactions on parallel and distributed systems, Vol. 16, No.11, pp 1066-1077.

Authors Profile

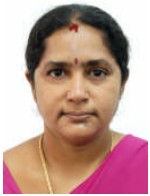

The author is a doctorate holder in Computer science with 21 years of experience in the academic side and approximately 2 years of experience in the industrial sector. She is the Professor and Head of the Department of Computer Science in Avinashilingam University for Women, Coimbatore-43. She has 80 publications at national and International level and executing funded projects worth 2 crores from UGC, AICTE and DRDO-NRB, DRDO-ARMREB. She is a life member of many professional organizations like CSI, ISTE, ISCA, WSEAS, AACE. Her areas of interest include network security, real time communication and real time operating systems. Her biography has been profiled at World's *Who's Who in Science and Engineering* Book, International Biographical Centre- Cambridge, England's - *Outstanding Scientist Worldwide for 2007, International Educator of the Year 2007* by IBC, *21st Century award for Achievement* by International Biographical Centre- Cambridge, England, *The International president's award for Iconic achievement*, by International Biographical Centre- Cambridge, England, *The Da Vinci Diamond for Inspirational Accomplishment* by International Biographical Centre, Cambridge.

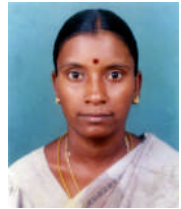

S.R.Vijayalakshmi is a Lecturer in School of Information Technology and Science, Dr.G.R.D college of science, Coimbatore. She received her B.Sc M.Sc, M.Phil in Electronics from the Bharathiar University and also received M.Sc in Computer Science from Bharathiar University and M.Phil in computer Science from Avinashilingam University for women. She has 14 years of teaching experience in the computer science and electronics field. Her research interests include embedded systems, parallel and distributed systems, real time systems, real time operating systems and microprocessors.